\pdfoutput=1

\documentclass[11pt]{article}

\usepackage{emnlp2021}

\usepackage{times}
\usepackage{latexsym}

\usepackage[T1]{fontenc}

\usepackage[utf8]{inputenc}

\usepackage{microtype}

\usepackage{array}
\newcolumntype{P}[1]{>{\centering\arraybackslash}p{#1}}

\usepackage{amsmath,amssymb,amsfonts}
\usepackage{graphicx}
\usepackage{color}
\usepackage{booktabs}
\usepackage{enumerate}
\usepackage{multicol, multirow}
\usepackage{arydshln}
\usepackage[english]{babel}
\usepackage[autostyle, english=american]{csquotes}
\MakeOuterQuote{"}
\usepackage{subcaption}
\usepackage{soul}
\usepackage{mwe}

\usepackage[ruled,vlined,linesnumbered]{algorithm2e}

\NewDocumentCommand{\tong}{ mO{} }{\textcolor{orange}{\textsuperscript{\textit{Tong}}\textsf{\textbf{\small[#1]}}}}
\NewDocumentCommand{\yingbo}{ mO{} }{\textcolor{purple}{\textsuperscript{\textit{Yingbo}}\textsf{\textbf{\small[#1]}}}}
\NewDocumentCommand{\semih}{ mO{} }{\textcolor{blue}{\textsuperscript{\textit{Semih}}\textsf{\textbf{\small[#1]}}}}

\title{Unsupervised Paraphrasing with Pretrained Language Models}

\author{
    Tong Niu \And Semih Yavuz \And Yingbo Zhou \AND Nitish Shirish Keskar \And Huan Wang \\ Salesforce Research \\ \texttt{\{tniu, syavuz, yingbo.zhou,} \\ \texttt{nkeskar, huan.wang, cxiong\}@salesforce.com} \And Caiming Xiong \\
}

\begin{document}
\maketitle

\begin{abstract}
Paraphrase generation has benefited extensively from recent progress in the designing of training objectives and model architectures. However, previous explorations have largely focused on supervised methods, which require a large amount of labeled data that is costly to collect. To address this drawback, we adopt a transfer learning approach and propose a training pipeline that enables pre-trained language models to generate high-quality paraphrases in an unsupervised setting. Our recipe consists of task-adaptation, self-supervision, and a novel decoding algorithm named Dynamic Blocking (DB). To enforce a surface form dissimilar from the input, whenever the language model emits a token contained in the source sequence, DB prevents the model from outputting the subsequent source token for the next generation step. We show with automatic and human evaluations that our approach achieves state-of-the-art performance on both the Quora Question Pair (QQP) and the ParaNMT datasets and is robust to domain shift between the two datasets of distinct distributions. We also demonstrate that our model transfers to paraphrasing in other languages without any additional finetuning.
\end{abstract}
\section{Introduction}
\label{sec:introduction}

\begin{figure}[t]
\centering
\includegraphics[width=0.48\textwidth]{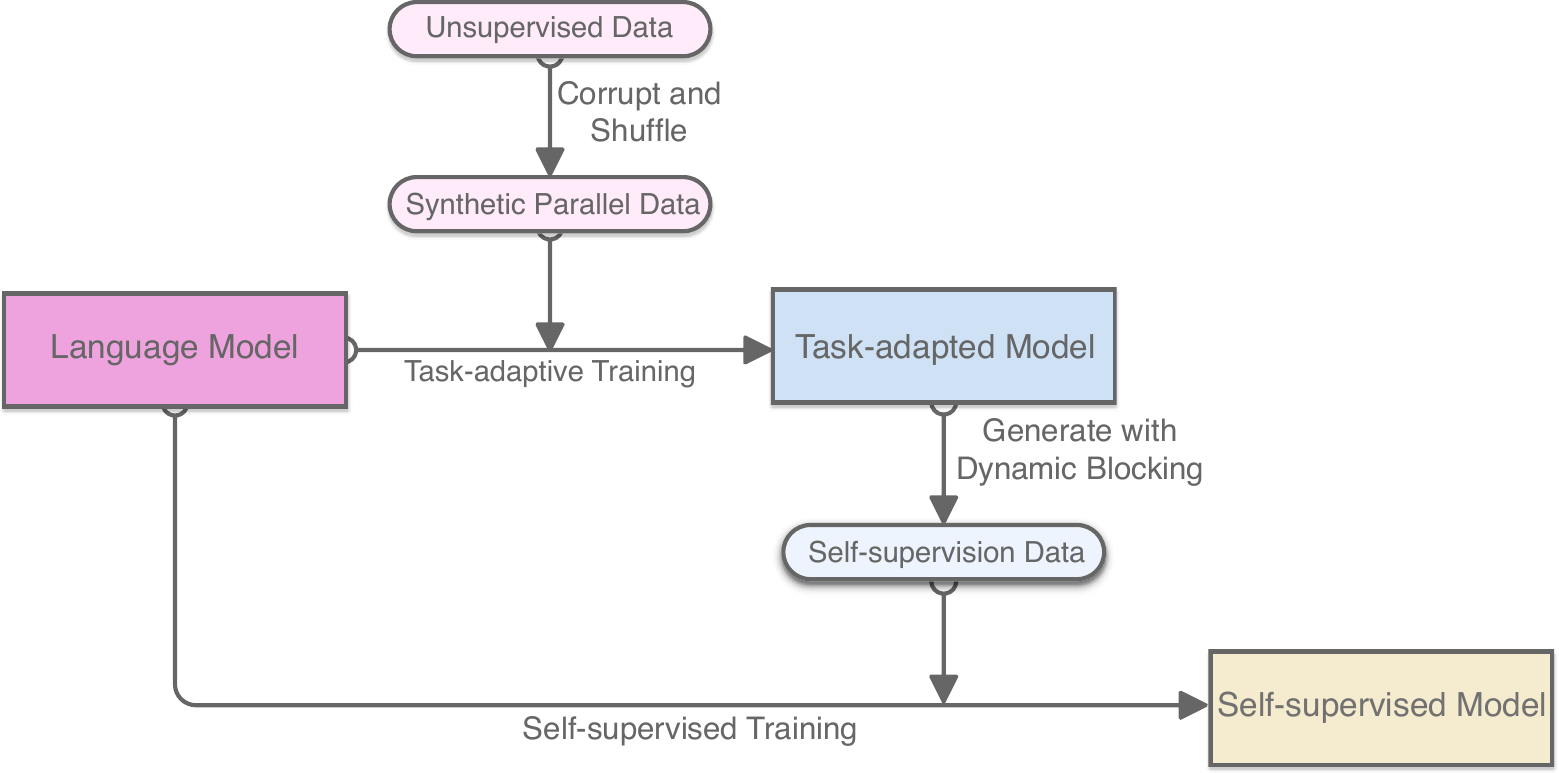}
\caption{Training pipeline of our paraphrasing model. We first train a task-adapted model with a denoising objective so that it is able to reconstruct input text. We then use Dynamic Blocking (DB) to generate pseudo-pairs of paraphrasing data. Finally, the generated data is used to train the self-supervised model.}
\label{fig:pipeline}
\end{figure}

Paraphrase generation restates text input in a different surface form while preserving its semantics. It has various applications on downstream NLP tasks including text summarization~\cite{cao2016joint}, semantic parsing~\cite{berant2014semantic}, as well as diversifying text generation for user-facing systems such as chatbots. To evaluate model robustness, a paraphraser can be used to generate adversarial examples, which also serve as augmented data to train the target neural networks~\cite{iyyer2018adversarial}. Besides, paraphrasing queries makes Question Answering systems more likely to match with keywords in a knowledge base~\cite{fader2014open, yin2015answering}.

However, it is expensive to annotate paraphrases, resulting in only a few human-labeled datasets. The existing ones are either small-scale like MRPC~\cite{dolan2005automatically}, or of closed domains like QQP\footnote{\url{https://www.kaggle.com/c/quora-question-pairs}} which consists entirely of questions. 
Consequently, previous work explored automatically (hence noisily) annotated datasets such as PIT-2015~\cite{xu2013gathering}, Twitter URL
Paraphrase Corpus~\cite{lan-etal-2017-continuously}, ParaNMT~\cite{wieting-gimpel-2018-paranmt}, and ParaBank~\cite{hu-etal-2019-large}, or re-purposed datasets including MSCOCO~\cite{lin2014microsoft} and WikiAnswers~\cite{fader2013paraphrase}. The scarcity of high-quality datasets motivates us to consider unsupervised alternatives. In this work, we explore a transfer learning approach, which leverages unsupervised large-scale pretrained models like T5~\cite{raffel2019exploring} and BART~\cite{lewis2019bart}. 

The effectiveness of BERT-\textit{score}~\cite{zhang2019bertscore} in identifying text similarity hints that pre-trained language models are 
equipped with extensive knowledge in paraphrasing. This knowledge may be attributed to the fact that text spans sharing similar context usually stay semantically close together -- word embedding~\cite{mikolov2013efficient} being a classic example. In other words, the paraphrasing capability of language models stems from 
the strong correlation between context and semantic similarity. In this work, we use pre-trained autoregressive LMs to leverage such implicit knowledge for paraphrasing in an unsupervised setting.\footnote{We will release all codes.}  

For paraphrasing, decoder-only LMs merely output a continuation of the input, while Sequence-to-Sequence models like BART tend to copy the input through even when paired with popular decoding algorithms such as greedy decoding, beam search or top-$k$/$p$ sampling~\cite{holtzman2019curious} because the probabilities of the input tokens during generation are all peaked. To address this issue, we propose \textit{Dynamic Blocking} (DB), a decoding algorithm that effortlessly transforms pre-trained autoregressive language models into natural paraphrasers with the help of \textit{task-adaption} and \textit{self-supervision} (Figure~\ref{fig:pipeline}). To obtain a surface form different from the input, whenever we emit a token that is present in the source sequence, this algorithm prevents the model from outputting its immediate successor for the next generation step. The algorithm is based on the intuition that during inference, although the top candidate at each generation step corresponds to a peaked probability, the rest of the distribution
still contains rich linguistic knowledge suitable for paraphrasing. This is in similar spirit with using soft targets for model distillation~\cite{hinton2015distilling}.

Through automatic and human evaluations, we demonstrate that our approach outperforms previous models (including supervised, in-domain models and the ground-truth targets) on both QQP and ParaNMT datasets and incurs no performance loss under domain shifts (i.e., finetuned on QQP and evaluated on ParaNMT, and vice versa). For automatic evaluations, we propose a \textit{reference-independent} automatic metric named BERT-\textit{i}BLEU, which is a harmonic mean of BERT-\textit{score} and one minus \textit{self}-BLEU. We show that this new metric correlates significantly better with human evaluation than traditional metrics. On the qualitative side, we illustrate with concrete examples that our model generates paraphrases that exhibit diverse syntactic structures. Finally, we observe that our model can generate paraphrases in other languages without any additional training.

Our contributions are: (1) a training pipeline that leads to a strong, unsupervised paraphrasing model; (2) a novel decoding algorithm that effectively diversifies paraphrase generation; (3) a new automatic metric that evaluates paraphrasing quality more accurately.

\section{Model}
\label{sec:model}
Figure~\ref{fig:pipeline} shows the training pipeline of our paraphrasing model, which consists of three key components, namely \textit{task-adaptation}, \textit{self-supervision} and \textit{Dynamic Blocking}. Overall we decode the task-adapted model with Dynamic Blocking to generate self-supervision data, which is in turn used to train the final model.

\begin{figure*}[t]
\centering
\includegraphics[width=1.0\textwidth]{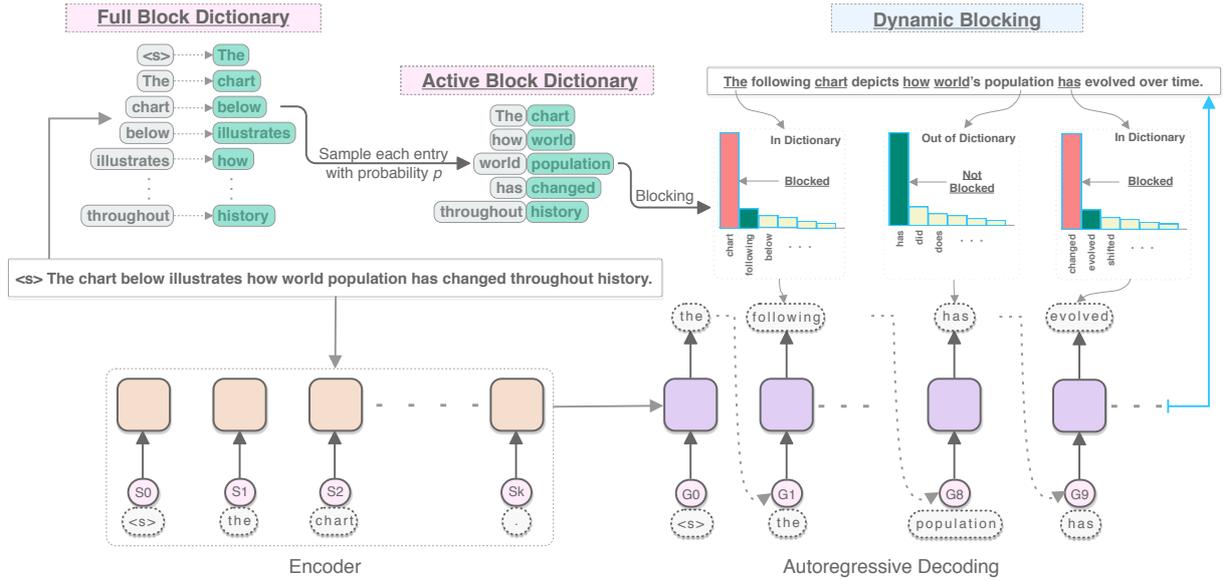}
\caption{Illustration of the Dynamic Blocking algorithm on real outputs. The algorithm first constructs a \textit{full block dictionary} based on the input, which maps each token to its immediate successor to be blocked, and then samples from this dictionary to build multiple \textit{active block dictionaries}, each used for generating a distinct paraphrase. When establishing an active dictionary, each entry in the full dictionary has a probability of $p$ to be sampled. During generation, the blocking takes place whenever an item in the active dictionary is triggered.}
\label{fig:dynamic blocking}
\end{figure*}

\subsection{Task-Adaptation}
\label{subsect:task-adaptation}
Inspired by~\newcite{gururangan-etal-2020-dont}, we apply task-adaptive training on the target dataset, treating its training set as a non-parallel collection of sentences. We perform task-adaptation by reconstructing the original sequence from its corrupted version with a denoising auto-encoder objective. Unlike previous work~\cite{devlin-etal-2019-bert,lewis2019bart}, we do not corrupt inputs with masks, but rather directly remove the corrupted tokens. This is to avoid pretrain-finetune discrepancy in denoising autoencoding models~\cite{yang2019xlnet}.
After the deletions, we randomly shuffle all remaining tokens to encourage the model to learn different alignments for better syntactic diversity.\footnote{For example, consider an input sentence "\textit{I want to lose weight in a healthy way.}" where we sample words "\textit{to}" and "\textit{way}" to delete and shuffle the rest. This may give us "\textit{weight in want a lose I healthy .}" as the corrupted sentence.} Note that we perform both deletions and shuffling on the word-level. This is similar to whole-word masking introduced in later versions of BERT~\cite{devlin-etal-2019-bert}. To demonstrate the benefit of our corruption strategy, we present ablation study results in Section~\ref{subsec:automatic-evaluation} by either adding masks or not shuffling. 

\subsection{Dynamic Blocking}
\label{subsec:dynamic-blocking}
Unlike previous diversity-promoting work which mainly focuses on the target side and encourages dissimilarity among beams~\cite{vijayakumar2018diverse,kumar-etal-2019-submodular,holtzman2019curious}, Dynamic Blocking takes the source input into account to guide the model toward generating in a different surface form (Figure~\ref{fig:dynamic blocking}).
As illustrated in Algorithm~\ref{alg:dynamic blocking}, we represent the source sequence $S$ as a list of tokens $S = (S_0, S_1, ..., S_M)$ and similarly the generated sequence as $G = (G_0, G_1, ..., G_N)$. Suppose that during generation, the model emits $G_j$ that is identical to some $S_i$ (it is not necessary that $i = j$). Then for the next generation step $G_{j+1}$, the algorithm forbids the model to generate $S_{i+1}$. 
Note that we block $S_{i+1}$ for only one step.
After $G_{j+1}$ is generated, we perform a different blocking for $G_{j+2}$ iff $G_{j+1} \in S$.

\begin{algorithm}[h!]
\small
\SetAlgoLined
\caption{Dynamic Blocking}
\label{alg:dynamic blocking}
\SetKwInOut{Input}{input}
\SetKwInOut{Output}{output}
\Input{A source sequence $S$ consisting of a list of tokens $S = (S_0, S_1, ..., S_M)$, and a $G_0 = \textsc{BOS}$ to start the decoding process}
Initialize $j \leftarrow 0$ \\
 \While{$G_j \neq \textsc{EOS}$ }{
  \If{$G_j = S_i \in S$ for some $i$}{
   $P(G_{j+1} = S_{i+1} | S, (G_0, G_1,..., G_j) \leftarrow 0$
   }
   Generate $G_{j+1}$ \\
   $j \leftarrow j + 1$ \\
 }
\Output{$G = (G_0, G_1, ..., G_N)$}
\end{algorithm}

The motivation to block for only one generation step is to allow the possibility of \textit{pure} syntactic variation of the original sequence, meaning that all tokens are kept but their order is permuted. For example, let us consider a decoding algorithm that completely prevents the model from generating a source token at all generation steps -- a popular n-gram blocking strategy we call \textit{Static Blocking}. Suppose that we intend to paraphrase "\textit{I like apples and oranges.}" as "\textit{I like oranges and apples.}". This is a valid paraphrase, but if we completely block the word "\textit{apples}" at all generation steps, it will be impossible to arrive at this paraphrase. However, with Dynamic Blocking the model will still be able to generate the word "\textit{apples}" later on even though this word has been temporarily blocked for one step after "\textit{and}" is generated. As shown in Figure~\ref{fig:dynamic blocking}, Dynamic Blocking builds a block dictionary which maps each token in the source sequence to its immediate successor. We then sample from this dictionary with a probability $p$ for each entry. This hyperparameter controls how different we want the paraphrase to be from the source input. In two extreme cases: when $p=0.0$, the model does not block any tokens and most likely copies through the source sequence; when $p = 1.0$, the model always blocks the immediate next token, leading to a drastically different surface form. In this work, we take the middle ground and set $p = 0.5$ so that for each blocking action, there will be half of the candidates taking that path. Note that if a word is tokenized into several subwords, \textit{only} the first subword is allowed to be blocked. 

We sample multiple block dictionaries to ensure \textit{diversity} among candidates, while leveraging beam search to ensure \textit{coherence}. For each sampled block dictionary, we use beam search to generate four candidates and keep the top-ranked two. It is beneficial to combine the two decoding methods because beam search helps to weed out ungrammatical or semantically invalid candidates.\footnote{For more details on Dynamic Blocking, please refer to Appendix~\ref{sec:details-of-dynamic-blocking}.}

Note that we only adopt bi-gram blocking because it is a superset of all higher-gram blockings. Consider, e.g., a tri-gram blocking entry $ab \rightarrow c$ in the block dictionary. If this entry is triggered, then the bi-gram blocking entry $b \rightarrow c$ will also have been triggered. Hence we found it unnecessary to include higher-order n-grams.


\subsection{Self-Supervision}
\label{subsect:self-supervision}
To help the model internalize patterns learned from task-adaption, 
we pseudo-label the training set~\cite{siddhant-etal-2020-leveraging} by decoding the task-adapted model with Dynamic Blocking. Having obtained the self-supervision data, we discard the task-adapted model and start from the pre-trained language model to avoid catastrophic forgetting~\cite{chronopoulou2019embarrassingly, chen-etal-2020-recall}. We also include reversed data (i.e., swapping source and target) because during task-adaptation the target is always longer than the input,
and including reversed data helps to offset this bias of sequence length. 


\section{Experimental Setup}
\label{sec:exp-setup}





\subsection{BERT-\textit{i}BLEU}
To evaluate paraphrasing quality, we propose a new metric named BERT-\textit{i}BLEU which encourages semantic closeness while penalizing surface-form similarity.
For semantic closeness we use the unsupervised metric BERT-\textit{score}~\cite{zhang2019bertscore}, which leverages a pretrained language model to compute the cosine similarity between each token in the candidate and that in the reference using contextual embeddings.\footnote{
In early experiments we tried another unsupervised metric Universal Sentence Encoder~\cite{cer-etal-2018-universal} and supervised metrics including RUSE~\cite{shimanaka-etal-2018-ruse}, Sentence-BERT~\cite{reimers-gurevych-2019-sentence}, and BLEURT~\cite{sellam-etal-2020-bleurt}. We observed that BERT-\textit{score} worked better at evaluating semantic similarity compared to these metrics.
}
To ensure that the key information (often conveyed through relatively rare words) is retained in the paraphrase, we apply IDF-reweighing on each token.\footnote{We use the BookCorpus dataset~\cite{Zhu_2015_ICCV} to compute the IDF weights.}
To measure the surface-form dissimilarity, we use one minus \textit{self}-BLEU, where \textit{self}-BLEU is the BLEU score between the source and the candidate. Hence BERT-\textit{i}BLEU (where \textit{i} stands for \textit{inverse}) is a weighted harmonic mean of the BERT-\textit{score} and one minus \textit{self}-BLEU.

\resizebox{0.96\linewidth}{!}{
  \begin{minipage}{\linewidth}
    \begin{align*}
        \mbox{BERT-\textit{i}BLEU} &= \left( \frac{\beta * \mbox{BERT-\textit{score}}^{-1} + 1.0 * (1 - \mbox{\textit{self}-BLEU})^{-1}}{\beta + 1.0} \right)^{-1} \\
        \mbox{\textit{self}-BLEU} &= \mbox{BLEU}(\mbox{\textit{source}}, \mbox{\textit{candidate}}) \\
    \end{align*}
  \end{minipage}
}
\\
As an extreme case, though copying through the input leads to a perfect BERT-\textit{score}, $1 - \mbox{\textit{self}-BLEU} = 0$; hence $\mbox{BERT-\textit{i}BLEU} = 0$. This is the reason that we do not use the BERT-\textit{score} directly to evaluate paraphrases. $\beta$ is used to control the relative importance between semantic similarity and surface-form dissimilarity. In our experiments we set $\beta = 4.0$ to scale up BERT-\textit{score} so that it has a similar range with \textit{self}-BLEU.
Note that because BERT-\textit{i}BLEU is reference-independent, it serves both as a metric to evaluate paraphrasing quality and as a criterion to re-rank generated candidates during task-adaptation and self-supervision.


\subsection{Dataset}
We evaluate on the Quora Question Pair (QQP) and the ParaNMT datasets.
QQP contains $140$K question pairs that are marked as a duplicate to each other and $640$K non-parallel questions. The sizes of dev and test sets are 3K and 20K, respectively. The ParaNMT dataset was constructed by back-translating sentences in Czech in the CzEng dataset~\cite{bojar2016czeng}. We directly obtained the test set of SOW-REAP from the authors of~\newcite{goyal-durrett-2020-neural}. To match the size of their training set, for task-adaptation we sample $350$K non-parallel sentences from ParaNMT-5M, while to generate self-supervision data we sample $350$K sentences from the same corpus as inputs. We filter out any sentences in SOW-REAP's test set to avoid training on test examples.

\subsection{Reproduction of Previous Models}
\label{subsec:reproduce-previous-models}
For the experiments on QQP we reproduce the supervised Transformer with the pre-trained T5-base model, which is stronger than the usual setting where the paraphraser trains from scratch. We also reproduce the model from~\newcite{hegde2020unsupervised}, which we refer to as CorruptLM. This model is similar to our task-adaptive phase (Section~\ref{subsect:task-adaptation}), except that they corrupt the inputs by removing all stop words rather than a fixed percentage of arbitrary words.\footnote{Because the original paper did not provide the source of the stop words, we extract the first $252$ words from The Corpus of Contemporary American English~\cite{davies2010corpus} to match the number.} Instead of GPT-2 as used by their work, we use BART which shows stronger results on downstream tasks. The rest of the settings remain the same.\footnote{To encourage the model to output new words in the reconstructed sentence, CorruptLM starts by randomly replacing $20\%$ of the words in the source sequence with synonyms using Syn-net~\cite{miller1998wordnet} (also applied during inference). 
} 
For the experiments on ParaNMT we use the SOW-REAP model released by~\newcite{goyal-durrett-2020-neural}.\footnote{\url{https://github.com/tagoyal/sow-reap-paraphrasing/}}

\subsection{Automatic Evaluation}
To evaluate paraphrasing quality, we follow~\newcite{li-etal-2019-decomposable} to report iBLEU~\cite{sun-zhou-2012-joint}, BLEU~\cite{papineni-etal-2002-bleu} and ROUGE~\cite{lin2004rouge} on QQP, and report BLEU and ROUGE on ParaNMT. Follwing~\newcite{goyal-durrett-2020-neural}, for ParaNMT both BLEU and ROUGE are calculated by first selecting the candidate that achieves the \textit{best} sentence-level score with the ground-truth, and then compute the corpus-level score of all these candidates. We use \textit{py-rouge}\footnote{\url{https://pypi.org/project/py-rouge/}} to compute ROUGE and the \textit{Datasets} library from \textit{HuggingFace}\footnote{\url{https://huggingface.co/metrics/sacrebleu}} to compute BLEU. We also report BERT-\textit{i}BLEU for the models we reproduced.

\subsection{Human Evaluation}
\label{subsec:human-evaluation}
We conduct human evaluations on MTurk.\footnote{Screenshots of the interfaces used by our MTurk studies are presented in Appendix~\ref{sec:mturk instructions}.} For each experiment, we compare our model with the strongest models reported in both supervised and unsupervised settings. On QQP, we compare with supervised Transformer, unsupervised CorruptLM, and the ground-truth. On ParaNMT, we compare with SOW-REAP and the ground-truth. To construct holistic human studies, we opt for both head-to-head binary comparison and Likert-scale scoring. The former provides straightforward results on which model is stronger, while the latter is used to consolidate their relative positions.

We only worked with annotators who had completed more than $10K$ assignments, had an approval rate of $>98\%$, and resided in the US. We also required that the annotators be native English speakers. When comparing between two model outputs based on the same input, we asked the annotators to identify which paraphrase they prefer in terms of overall quality.\footnote{We intentionally did not ask them to separately evaluate semantic similarity and surface-form diversity because the latter is easy to check with \textit{self}-BLEU.} For each experiment, we randomly sampled $200$ examples from the QQP's or ParaNMT's test set and shuffled the order of each example to anonymize the model identities. Each assignment was scored by two annotators.
\section{Results}
\label{sec:results}

\subsection{Human Evaluation}
Table~\ref{tab:human-binary} and~\ref{tab:human-likert} present human evaluation results on our final model compared with other baselines.
On QQP our model outperforms both Transformer and CorruptLM. Recall that CorruptLM also leverages a pre-trained language model. This indicates the effectiveness of our 
training pipeline
when holding the LM factor as a constant. On ParaNMT our model outperforms SOW-REAP in both head-to-head and Likert-based evaluations. Moreover, our model outperforms the ground-truth on both datasets. For ParaNMT, the result indicates that our approach also outperforms a supervised round-trip translation baseline since that is how ParaNMT data was generated in the first place. For QQP, we note two reasons why these scores do not indicate that our model can generate paraphrases with human-level quality. First, QQP is human-labeled, not human-generated.
Second, QQP annotates duplicate questions rather than paraphrases. Questions referring to the same topic but are not semantically equivalent may still be marked as duplicates.\footnote{For instance, the question pair "\textit{I'm 27, is it too late for me to go to medical school?}" and "\textit{How old is too old to start medical school?}" has a positive label even though they do not share the same meaning.}

\begin{table}[t]
\centering
\small
\setlength{\tabcolsep}{2pt}
    \begin{tabular}{cccccc}
    \toprule
    \multicolumn{1}{c}{Dataset} & \multicolumn{1}{l}{\textbf{Ours} v.s.} & Win($\%$) & Tie($\%$) & Loss($\%$) & W-L($\%$) \\
    \midrule
    \multirow{3}[0]{*}{QQP}
        & Transformer  & $40.75$ & $28.25$ & $31.00$ & $12.50$ \\
        & CorruptLM    & $46.00$ & $26.25$ & $27.75$ & $18.00$ \\
        & Ground-truth & $43.00$ & $16.75$ & $40.25$ & $\;2.75$ \\
    \midrule
    \multirow{2}[0]{*}{ParaNMT}
        & SOW-REAP     & $40.50$ & $28.50$ & $31.00$ & $\;9.50$ \\
        & Ground-truth & $49.50$ & $14.50$ & $36.00$ & $13.50$ \\
    \bottomrule
    \end{tabular}%
  \caption{Head-to-head human evaluation results. Each experiment is performed over 200 samples with 2 annotators each. 
  "\textbf{Ours}" stands for the model trained with self-supervision and decoded with Dynamic Blocking. Note that both Transformer and SOW-REAP are supervised models, and we are also comparing our unsupervised model outputs with the ground-truth. "\textbf{W-L}" stands for the difference between \textit{Win} and \textit{Loss}.}
  \label{tab:human-binary}%
\end{table}%

We use Cohen's Kappa to evaluate the inter-annotator agreement. For head-to-head evaluations, we obtained \textit{kappa} $= 0.35$, indicating fair agreement. Note that when calculating kappa, we leave out all cases where either of the two annotators gives a "tie" because this usually signifies that they are unsure about which paraphrase is better.

    
\begin{table}[t]
\centering
\small
    \begin{tabular}{cccc}
    \toprule
    Dataset  & \multicolumn{2}{c}{Model} & Avg. Score\\
    \midrule
    \multirow{3}[0]{*}{QQP} & Supervised & Transformer & $4.04 \pm 1.01$  \\
    \cmidrule{2-3} & \multirow{2}[0]{*}{Unsupervised} & CorruptLM & $3.74 \pm 1.26$ \\
        &  & \textbf{Ours} & $\textbf{4.19} \pm \textbf{0.99}$  \\
    \midrule
    \multirow{2}[0]{*}{ParaNMT} & \multirow{1}[0]{*}{Supervised} 
                 & SOW-REAP & $3.78 \pm 1.15$    \\
\cmidrule{2-3}          & Unsupervised & \textbf{Ours} & $\textbf{3.94} \pm \textbf{1.09}$    \\
    \bottomrule
    \end{tabular}%
  \caption{Likert-scale human evaluation results. Both averages and standard deviations are reported.}
  \label{tab:human-likert}%
\end{table}%

\subsection{Advantage of the Proposed Metric}
\label{subsec:curse-of-bleu-on-paraphrase-evaluation}
To facilitate a better understanding of the automatic evaluation results, we investigate how each of the automatic metrics correlates with human evaluation. Table~\ref{tab:correlation} shows that BERT-\textit{i}BLEU agrees significantly better with human perceptions. The reason that BLEU does not correlate well with human evaluation is that there are two conflicting objectives. The first comes from keeping the important information, such as named entities, which should be copied verbatim, while the second comes from using different wordings to express the same semantics -- the better the model is at this,
the lower the BLEU becomes. For a model good at both, the gain in BLEU for matching key entities and the loss for using different wordings cancel each other out, preventing BLEU from faithfully evaluating the paraphrasing quality. Consequently, BLEU is only useful for checking extreme cases: very low or high BLEU usually signals bad paraphrases, but for the middle-ground cases BLEU alone is less indicative. A similar argument holds for ROUGE. In contrast, BERT-\textit{score} encourages the first objective and is not penalized by the second.
However, parroting the input will still fool BERT-\textit{score} alone. Hence we pair it with \textit{self}-BLEU to encourage surface-form diversity.

\begin{table}[t]
\small
  \centering
    \begin{tabular}{ccccc}
    \toprule
          & BERT-\textit{i}BLEU & iBLEU & BLEU & ROUGE-1/2/L \\
    \midrule
    Agree $\%$ & \textbf{68.9} & 39.4 & 45.3 & 21.8/5.4/21.4 \\
    \bottomrule
    \end{tabular}%
  \caption{The percentage of times where the ranking given by each metric agrees with that given by human evaluation in the head-to-head studies. Only cases where two annotators agree are counted.}
  \label{tab:correlation}%
\end{table}%

\begin{table*}[t]
\centering
\setlength{\tabcolsep}{2.75pt}
  \centering
    \begin{tabular}{llcccccc}
    \toprule
        \multicolumn{2}{c}{Model} & BERT-\textit{i}BLEU & iBLEU & BLEU & ROUGE-1 & ROUGE-2 & ROUGE-L \\
    \midrule
    \multirow{1}[0]{*}{Supervised} 
        & Transformer & 68.7 & 17.0 & 22.3 & 55.8 & 32.3 & 57.5 \\
    \midrule
    \multirow{5}[0]{*}{Unsupervised} 
        & CorruptLM & 61.5 & 12.1 & 16.8 & 49.1 & 26.2 & 51.7 \\
        & TA  & 76.2  & 16.0 & 21.2 & 61.9 & 35.1 & 61.7 \\
        & TA+SS & 78.9 & 15.6 & 20.7 & 61.5 & 32.8 & 60.7 \\
        & TA+SS+DB (NMT) & 82.5 & 10.1 & 14.6 & 60.1 & 28.5 & 58.6 \\
        & \textbf{TA+SS+DB} & \textbf{83.1} & 9.6 & 14.1 & 59.9 & 28.5 & 58.8 \\
    \midrule
    No Model & Copy-input & 0.0 & \textbf{24.3} & \textbf{30.4} & \textbf{65.7} & \textbf{41.7} & \textbf{66.5} \\
    \bottomrule
    \end{tabular}%
  \caption{
    Automatic evaluation results on QQP. TA = \textbf{T}ask-\textbf{A}daptation, SS = \textbf{S}elf-\textbf{S}upervision and DB = \textbf{D}ynamic \textbf{B}locking. "NMT" stands for model finetuned on non-parallel ParaNMT and evaluated cross-domain on QQP. Both our final model (TA+SS+DB) and the best result for each metric are boldfaced. Please refer to Section~\ref{sec:automatic-metric-results} in the Appendix for a comparison with $12$ supervised models and $5$ unsupervised models from previous work.
  }
  \label{tab:qqp}%
\end{table*}%

\begin{table*}[t]
  \centering
    \begin{tabular}{llccccc}
    \toprule
    \multicolumn{2}{c}{Model} & BERT-\textit{i}BLEU & BLEU & ROUGE-1 & ROUGE-2 & ROUGE-L \\
    \midrule
    \multirow{1}[0]{*}{Supervised} 
        & SOW-REAP & 54.2 & \textbf{30.9} & \textbf{62.3} & \textbf{40.2} & \textbf{61.7} \\
    \midrule
    \multirow{5}[0]{*}{Unsupervised}
        & CorruptLM (QQP) & 39.7 & 7.6 & 31.9 & 11.6 & 31.6 \\
        & TA & 72.0 & 20.2 & 59.0 & 32.3 & 53.8 \\
        & TA+SS & 74.0 & 22.9 & 58.9 & 33.3 & 54.1 \\
        & TA+SS+DB (QQP) & 76.8 & 22.0 & 60.1 & 33.8 & 54.9 \\
        & \textbf{TA+SS+DB} & \textbf{78.0} & 22.6 & 59.8 & 33.2 & 54.5\\
    \midrule
    \multirow{1}[0]{*}{No Model}
        & Copy-input & 0.0 & 18.4 & 54.4 & 27.2 & 49.2 \\
    \bottomrule
    \end{tabular}%
  \caption{Automatic evaluation results on ParaNMT. "QQP" stands for models finetuned on non-parallel QQP and evaluated cross-domain on ParaNMT. Note that BLEU and ROUGE scores are based on top-$10$ candidates where \textbf{only the ones with the highest sentence-level scores} are retained for the final score computation.
  }
  \label{tab:nmt}%
\end{table*}%

\subsection{Automatic Evaluation}
\label{subsec:automatic-evaluation}
On QQP, our model outperforms both the supervised Transformer and the unsupervised CorruptLM on BERT-\textit{i}BLEU (Table~\ref{tab:qqp}).\footnote{
We tried combining supervised Transformer with DB, and obtained a BERT-iBLEU of $80.1$ on QQP, indicating that DB itself is an effective diversity-promoting decoding strategy.
}
Recall that both Transformer and CorruptLM leverage a strong pretrained language model, indicating that the performance gain stems mainly from our proposed pipeline rather than the language model itself. 
On ParaNMT, our model outperforms the supervised SOW-REAP (Table~\ref{tab:nmt}).\footnote{Please refer to Appendix~\ref{sec:automatic-metric-results} for results of our model compared with all previous ones on the traditional metrics.} As ablation studies on task-adaptation and self-supervision, we can see in Table~\ref{tab:qqp} and~\ref{tab:nmt} that our model (TA+SS+DB) beats the one that is either task-adapted only (TA) or self-supervised but decoded without DB (TA+SS), showing that both self-supervision and Dynamic Blocking are crucial to paraphrasing quality. 

On the traditional metrics in Table~\ref{tab:qqp}, our models also obtain competitive results with the supervised models. However, as we move down to the last row, we see that Copy-input achieves state-of-the-art results on all metrics except BERT-\textit{i}BLEU, indicating that iBLEU, BLEU, and ROUGE scores are not reliable for evaluating paraphrasing quality.\footnote{\newcite{mao-lee-2019-polly} also observe that parroting often achieves competitive results.} In contrast, our best model on BERT-\textit{i}BLEU (TA+SS+DB) achieves much lower iBLEU and BLEU scores as compared to other models, showing the inconsistency between these traditional metrics and human evaluation. We also note one special aspect of Table~\ref{tab:nmt} to make it easier to interpret. Unlike on QQP, the performance of Copy-input on ParaNMT is the lowest among all models. However, we need to take this comparison with a grain of salt because all the other results are based on $10$ candidates where only the ones with the highest sentence-level scores are retained. In contrast, Copy-input only has one candidate. Thus Copy-input and the other results are not directly comparable. Plus, SOW-REAP filters the dataset to only include syntactically diverse targets and then splits it into the train, dev and test sets, which makes Copy-input less effective.


\subsection{Robustness to Domain Shift}
On the ParaNMT dataset, we notice that CorruptLM, when finetuned on non-parallel QQP, achieves much worse results than the other models (CorruptLM (QQP) row in Table~\ref{tab:nmt}), indicating that it is less robust to domain shift. In contrast, our model achieves similar results compared to the in-domain one under the same setting (TA+SS+DB (QQP) row). Conversely, we also finetune our model on non-parallel ParaNMT and evaluate on QQP (TA+SS+DB (ParaNMT) row in Table~\ref{tab:qqp}). We observe that this model again achieves performance similar to that of the in-domain model. These results show that our model may be able to perform task-adaptation using an arbitrary out-of-domain corpus 
and still work well on the target domain.

\subsection{Ablation Studies on Corruption Strategies}
During task-adaptation, our corruption strategies involve both deletions and shuffling. In Table~\ref{tab:ablation} we provide ablation study results where we either replace words with masks instead of deleting them or delete words without shuffling. We can see that our delete-and-shuffle strategy achieves the best BERT-\textit{i}BLEU score among the three settings.

\begin{table}[t]
\centering
\small
  \begin{tabular}{cccc}
    \toprule
    & AddMask & NoShuffle & Delete-Shuffle \\
    \midrule
    BERT-\textit{i}BLEU & 80.7 & 81.7 & \textbf{83.1} \\
    \bottomrule
  \end{tabular}%
  \caption{Ablation studies on different corruption strategies for task-adaptation on QQP. AddMask stands for the strategy where corrupted words are replaced with MASK tokens; NoShuffle corresponds to "no shuffling" after sentence corruption.}
  \label{tab:ablation}%
\end{table}%
\section{Analysis}
\subsection{Syntactic Diversity}
In Table~\ref{tab:diversity}, we qualitatively demonstrate paraphrases generated by our model that exhibit syntactic structure variance. Unlike previous work relying on explicit syntactic scaffolding~\cite{goyal-durrett-2020-neural}, our model achieves syntactic diversity "for free" from shuffling during task-adaptation.\footnote{We present in Appendix~\ref{sec:robustness-to-grammar-errors} that shuffling also makes the model robust to grammar errors, enabling it to paraphrase and perform text normalization at the same time.}


\begin{table*}[t]
\small
  \centering
    \begin{tabular}{ll}
    \toprule
    \multicolumn{1}{c}{Input} & \multicolumn{1}{c}{Generated paraphrase} \\
    \midrule
    We got to spend the rest of the weekend at the track. yeah. & We got to stay at the track for the rest of the weekend. yeah. \\
    Are predictions of the future based on the present too much? & Are future predictions too much based on the present? \\
    What is the best way to reduce belly and arm fat?   &    What is the easiest way to reduce arm and belly fat? \\
    You can seduce enemy soldiers, though. & You can, though, seduce enemy troops. \\
    Well, why would your buddy be in the shower with you?! &  Okay, why would you be in the shower with your friend?! \\
    \bottomrule
    \end{tabular}%
  \caption{Selected paraphrases generated by our final model that shows syntactic variance at different extents. Only the top candidate is shown for each input.}
  \label{tab:diversity}%
\end{table*}%

\subsection{Generalization to Other Languages}
\label{subsec:generalization-to-other-languages}
\paragraph{Dynamic Blocking on BART without Finetuning}
\label{para:dynamic-blocking-without-finetuning}
Though we focus on T5 throughout the paper,
we do note a \textit{unique ability} of BART: it can directly work with Dynamic Blocking to generate paraphrases (i.e., without domain-adaptation and self-supervision), though of lower quality than the self-supervised model. We demonstrate such examples in Appendix~\ref{sec:block-inflections}.

\paragraph{Paraphrasing in Other Languages}
We observe that although BART is trained almost exclusively on English text, it is able to paraphrase in multiple other languages. We adopt the aforementioned BART setting and present an example in German (Table~\ref{tab:de} in Appendix~\ref{sec:paraphrasing-in-german}). To our best knowledge, this is the first unsupervised model that can paraphrase in a non-English language. 
The reasoning behind this observation is twofold. First, although BART was trained on English corpora, there is a small portion of the content in German due to mislabeled language identification, allowing the model to observe German data; second, previous work has shown that large-scale language models are able to perform zero-shot cross-lingual transfer on a variety of downstream classification tasks, such as Named Entity Recognition~\cite{moon2019towards}, Natural Language Inference, and Document Classification~\cite{artetxe-schwenk-2019-massively}. Our work hence demonstrates that it is possible to perform such a transfer even for generative tasks like paraphrasing.
We also hypothesize that the paraphrasing quality should improve if we apply our training pipeline to mBART or mT5~\cite{xue2020mt5}. We leave this as future work.
\section{Related Work}
\label{sec:related work}
Paraphrase generation has been a long-standing task that has several applications on downstream NLP tasks including text summarization \citep{cao2016joint}, semantic parsing \citep{berant2014semantic}, and question answering \citep{yu2018qanet}. 
Early works on paraphrase generation mostly rely on rule-based or statistical machine translation systems~\cite{mckeown1980paraphrasing,meteer1988strategies,bannard2005paraphrasing}. 

\paragraph{Supervised Approaches}
Neural sequence-to-sequence (Seq2Seq) models have been used to address this task~\cite{prakash-etal-2016-neural,li2017paraphrase, see2017pointergenerator, vaswani2017attention, gupta2018vae}; sometimes such models are also used to evaluate paraphrasing quality~\cite{thompson-post-2020-automatic}. Round-trip translation between two languages (i.e., back-translation) with strong neural machine translation (NMT) models has also become a widely used approach for paraphrase generation~\cite{yu2018qanet}. Consequently, supervised models using datasets like ParaNMT obtain their performance mainly from sequence-level distillation~\cite{kim-rush-2016-sequence}, where the data comes from the underlying supervised translation models.
There have been several previous works~\cite{iyyer-etal-2018-adversarial, chen2019multitask, li-etal-2019-decomposable,kumar-etal-2019-submodular,goyal-durrett-2020-neural} that make use of syntactic structures to produce more diverse paraphrases.
More recently,~\newcite{qian2019exploring} employ distinct generators to produce diverse paraphrases. 
Retrieval-augmented generation methods have also been investigated~\cite{kazemnejad2020paraphrase, lewis-etal-2020-bart}.
However, most of these approaches require parallel data.

\paragraph{Unsupervised Approaches}
Unsupervised paraphrasing, on the other hand, is a rather less explored and more challenging problem in NLP.~\newcite{bowman2016generating} train a variational autoencoder (VAE) to maximize the lower bounds for the reconstruction log-likelihood of the input sentence without requiring any parallel corpora. Sampling from the trained VAE's decoder leads to sentences that can practically be considered as paraphrases as the decoder aims to reconstruct the input sentence by its training objective.~\newcite{miao2018cgmh} introduce a constrained sentence generation approach by using Metropolis-Hastings sampling, which allows for decoding with complicated discrete constraints such as the occurrence of multiple keywords, hence not requiring any parallel corpora.~\newcite{roy-grangier-2019-unsupervised} introduce a model that allows interpolation from continuous auto-encoders to vector-quantized auto-encoders.~\newcite{liu-etal-2020-unsupervised} cast the paraphrasing as an optimization problem, where it searches the sentence space to find the optimal point for an objective function that takes semantic similarity, expression diversity, and language fluency into account.~\newcite{siddique2020unsupervised} optimize a similar objective with deep reinforcement learning.

\paragraph{Transfer Learning}
There have been few works leveraging pre-trained language models for paraphrasing, either in a supervised ~\cite{witteveen2019paraphrasing} or an unsupervised~\cite{hegde2020unsupervised} setting. Both works employ GPT-2 as their backbone generation model. Similarly, we opt for more recent large-scale pre-trained models like BART and T5.
\section{Conclusion}
We design an effective training pipeline that enables large-scale pre-trained models to generate high-quality paraphrases in an unsupervised setting through task-adaptation, self-supervision, and a novel decoding algorithm named Dynamic Blocking. We demonstrate with automatic and human evaluations that our model achieves state-of-the-art results on benchmark datasets. We also show that our model generates paraphrases that exhibit syntactic diversity, as well as generalizes to other languages without any additional training. 
Overall our work motivates a deeper investigation into self-supervised techniques for paraphrase generation as well as extensions such as \textit{context-aware} paraphrasing, where the output conditions not only on the sentences to be paraphrased, but also on the context around them. We leave this as future work.

\bibliography{anthology,custom}

\begin{thebibliography}{66}
\expandafter\ifx\csname natexlab\endcsname\relax\def\natexlab#1{#1}\fi

\bibitem[{Artetxe and Schwenk(2019)}]{artetxe-schwenk-2019-massively}
Mikel Artetxe and Holger Schwenk. 2019.
\newblock \href {https://doi.org/10.1162/tacl_a_00288} {Massively multilingual
  sentence embeddings for zero-shot cross-lingual transfer and beyond}.
\newblock \emph{Transactions of the Association for Computational Linguistics},
  7:597--610.

\bibitem[{Bannard and Callison-Burch(2005)}]{bannard2005paraphrasing}
Colin Bannard and Chris Callison-Burch. 2005.
\newblock Paraphrasing with bilingual parallel corpora.
\newblock In \emph{Proceedings of the 43rd Annual Meeting of the Association
  for Computational Linguistics (ACL’05)}, pages 597--604.

\bibitem[{Berant and Liang(2014)}]{berant2014semantic}
Jonathan Berant and Percy Liang. 2014.
\newblock Semantic parsing via paraphrasing.
\newblock In \emph{Proceedings of the 52nd Annual Meeting of the Association
  for Computational Linguistics (Volume 1: Long Papers)}, pages 1415--1425.

\bibitem[{Bojar et~al.(2016)Bojar, Du{\v{s}}ek, Kocmi, Libovick{\`y},
  Nov{\'a}k, Popel, Sudarikov, and Vari{\v{s}}}]{bojar2016czeng}
Ond{\v{r}}ej Bojar, Ond{\v{r}}ej Du{\v{s}}ek, Tom Kocmi, Jind{\v{r}}ich
  Libovick{\`y}, Michal Nov{\'a}k, Martin Popel, Roman Sudarikov, and
  Du{\v{s}}an Vari{\v{s}}. 2016.
\newblock Czeng 1.6: enlarged czech-english parallel corpus with processing
  tools dockered.
\newblock In \emph{International Conference on Text, Speech, and Dialogue},
  pages 231--238. Springer.

\bibitem[{Bowman et~al.(2016)Bowman, Vilnis, Vinyals, Dai, Jozefowicz, and
  Bengio}]{bowman2016generating}
Samuel~R. Bowman, Luke Vilnis, Oriol Vinyals, Andrew Dai, Rafal Jozefowicz, and
  Samy Bengio. 2016.
\newblock Generating sentences from a continuous space.
\newblock In \emph{Proceedings of The 20th {SIGNLL} Conference on Computational
  Natural Language Learning}.

\bibitem[{Cao et~al.(2016)Cao, Luo, Li, and Li}]{cao2016joint}
Ziqiang Cao, Chuwei Luo, Wenjie Li, and Sujian Li. 2016.
\newblock Joint copying and restricted generation for paraphrase.
\newblock \emph{arXiv preprint arXiv:1611.09235}.

\bibitem[{Cer et~al.(2018)Cer, Yang, Kong, Hua, Limtiaco, St.~John, Constant,
  Guajardo-Cespedes, Yuan, Tar, Strope, and Kurzweil}]{cer-etal-2018-universal}
Daniel Cer, Yinfei Yang, Sheng-yi Kong, Nan Hua, Nicole Limtiaco, Rhomni
  St.~John, Noah Constant, Mario Guajardo-Cespedes, Steve Yuan, Chris Tar,
  Brian Strope, and Ray Kurzweil. 2018.
\newblock \href {https://doi.org/10.18653/v1/D18-2029} {Universal sentence
  encoder for {E}nglish}.
\newblock In \emph{Proceedings of the 2018 Conference on Empirical Methods in
  Natural Language Processing: System Demonstrations}, pages 169--174,
  Brussels, Belgium. Association for Computational Linguistics.

\bibitem[{Chen et~al.(2019)Chen, Tang, Wiseman, and Gimpel}]{chen2019multitask}
Mingda Chen, Qingming Tang, Sam Wiseman, and Kevin Gimpel. 2019.
\newblock A multi-task approach for disentangling syntax and semantics in
  sentence representations.
\newblock In \emph{Proceedings of the 2019 Conference of the North {A}merican
  Chapter of the Association for Computational Linguistics: Human Language
  Technologies, Volume 1 (Long and Short Papers)}.

\bibitem[{Chen et~al.(2020)Chen, Hou, Cui, Che, Liu, and
  Yu}]{chen-etal-2020-recall}
Sanyuan Chen, Yutai Hou, Yiming Cui, Wanxiang Che, Ting Liu, and Xiangzhan Yu.
  2020.
\newblock \href {https://doi.org/10.18653/v1/2020.emnlp-main.634} {Recall and
  learn: Fine-tuning deep pretrained language models with less forgetting}.
\newblock In \emph{Proceedings of the 2020 Conference on Empirical Methods in
  Natural Language Processing (EMNLP)}, pages 7870--7881, Online. Association
  for Computational Linguistics.

\bibitem[{Chronopoulou et~al.(2019)Chronopoulou, Baziotis, and
  Potamianos}]{chronopoulou2019embarrassingly}
Alexandra Chronopoulou, Christos Baziotis, and Alexandros Potamianos. 2019.
\newblock An embarrassingly simple approach for transfer learning from
  pretrained language models.
\newblock In \emph{Proceedings of the 2019 Conference of the North {A}merican
  Chapter of the Association for Computational Linguistics: Human Language
  Technologies, Volume 1 (Long and Short Papers)}.

\bibitem[{Davies(2010)}]{davies2010corpus}
Mark Davies. 2010.
\newblock The corpus of contemporary american english as the first reliable
  monitor corpus of english.
\newblock \emph{Literary and linguistic computing}, 25(4):447--464.

\bibitem[{Devlin et~al.(2019)Devlin, Chang, Lee, and
  Toutanova}]{devlin-etal-2019-bert}
Jacob Devlin, Ming-Wei Chang, Kenton Lee, and Kristina Toutanova. 2019.
\newblock \href {https://doi.org/10.18653/v1/N19-1423} {{BERT}: Pre-training of
  deep bidirectional transformers for language understanding}.
\newblock In \emph{Proceedings of the 2019 Conference of the North {A}merican
  Chapter of the Association for Computational Linguistics: Human Language
  Technologies, Volume 1 (Long and Short Papers)}, pages 4171--4186,
  Minneapolis, Minnesota. Association for Computational Linguistics.

\bibitem[{Dolan and Brockett(2005)}]{dolan2005automatically}
William~B Dolan and Chris Brockett. 2005.
\newblock Automatically constructing a corpus of sentential paraphrases.
\newblock In \emph{Proceedings of the Third International Workshop on
  Paraphrasing (IWP2005)}.

\bibitem[{Fader et~al.(2013)Fader, Zettlemoyer, and
  Etzioni}]{fader2013paraphrase}
Anthony Fader, Luke Zettlemoyer, and Oren Etzioni. 2013.
\newblock Paraphrase-driven learning for open question answering.
\newblock In \emph{Proceedings of the 51st Annual Meeting of the Association
  for Computational Linguistics (Volume 1: Long Papers)}, pages 1608--1618.

\bibitem[{Fader et~al.(2014)Fader, Zettlemoyer, and Etzioni}]{fader2014open}
Anthony Fader, Luke Zettlemoyer, and Oren Etzioni. 2014.
\newblock Open question answering over curated and extracted knowledge bases.
\newblock In \emph{Proceedings of the 20th ACM SIGKDD international conference
  on Knowledge discovery and data mining}, pages 1156--1165.

\bibitem[{Goyal and Durrett(2020)}]{goyal-durrett-2020-neural}
Tanya Goyal and Greg Durrett. 2020.
\newblock \href {https://doi.org/10.18653/v1/2020.acl-main.22} {Neural
  syntactic preordering for controlled paraphrase generation}.
\newblock In \emph{Proceedings of the 58th Annual Meeting of the Association
  for Computational Linguistics}, pages 238--252, Online. Association for
  Computational Linguistics.

\bibitem[{Gupta et~al.(2018)Gupta, Agarwal, Singh, and Rai}]{gupta2018vae}
Ankush Gupta, Arvind Agarwal, Prawaan Singh, and Piyush Rai. 2018.
\newblock A deep generative framework for paraphrase generation.
\newblock In \emph{AAAI Conference on Artificial Intelligence}.

\bibitem[{Gururangan et~al.(2020)Gururangan, Marasovi{\'c}, Swayamdipta, Lo,
  Beltagy, Downey, and Smith}]{gururangan-etal-2020-dont}
Suchin Gururangan, Ana Marasovi{\'c}, Swabha Swayamdipta, Kyle Lo, Iz~Beltagy,
  Doug Downey, and Noah~A. Smith. 2020.
\newblock \href {https://doi.org/10.18653/v1/2020.acl-main.740} {Don{'}t stop
  pretraining: Adapt language models to domains and tasks}.
\newblock In \emph{Proceedings of the 58th Annual Meeting of the Association
  for Computational Linguistics}, pages 8342--8360, Online. Association for
  Computational Linguistics.

\bibitem[{Hegde and Patil(2020)}]{hegde2020unsupervised}
Chaitra Hegde and Shrikumar Patil. 2020.
\newblock Unsupervised paraphrase generation using pre-trained language models.
\newblock \emph{arXiv preprint arXiv:2006.05477}.

\bibitem[{Hinton et~al.(2015)Hinton, Vinyals, and Dean}]{hinton2015distilling}
Geoffrey Hinton, Oriol Vinyals, and Jeff Dean. 2015.
\newblock Distilling the knowledge in a neural network.
\newblock \emph{arXiv preprint arXiv:1503.02531}.

\bibitem[{Holtzman et~al.(2020)Holtzman, Buys, Du, Forbes, and
  Choi}]{holtzman2019curious}
Ari Holtzman, Jan Buys, Li~Du, Maxwell Forbes, and Yejin Choi. 2020.
\newblock The curious case of neural text degeneration.
\newblock In \emph{The Ninth International Conference on Learning
  Representations}.

\bibitem[{Hu et~al.(2019)Hu, Singh, Holzenberger, Post, and
  Van~Durme}]{hu-etal-2019-large}
J.~Edward Hu, Abhinav Singh, Nils Holzenberger, Matt Post, and Benjamin
  Van~Durme. 2019.
\newblock \href {https://doi.org/10.18653/v1/K19-1005} {Large-scale, diverse,
  paraphrastic bitexts via sampling and clustering}.
\newblock In \emph{Proceedings of the 23rd Conference on Computational Natural
  Language Learning (CoNLL)}, pages 44--54, Hong Kong, China. Association for
  Computational Linguistics.

\bibitem[{Iyyer et~al.(2018{\natexlab{a}})Iyyer, Wieting, Gimpel, and
  Zettlemoyer}]{iyyer2018adversarial}
Mohit Iyyer, John Wieting, Kevin Gimpel, and Luke Zettlemoyer.
  2018{\natexlab{a}}.
\newblock Adversarial example generation with syntactically controlled
  paraphrase networks.
\newblock \emph{arXiv preprint arXiv:1804.06059}.

\bibitem[{Iyyer et~al.(2018{\natexlab{b}})Iyyer, Wieting, Gimpel, and
  Zettlemoyer}]{iyyer-etal-2018-adversarial}
Mohit Iyyer, John Wieting, Kevin Gimpel, and Luke Zettlemoyer.
  2018{\natexlab{b}}.
\newblock \href {https://doi.org/10.18653/v1/N18-1170} {Adversarial example
  generation with syntactically controlled paraphrase networks}.
\newblock In \emph{Proceedings of the 2018 Conference of the North {A}merican
  Chapter of the Association for Computational Linguistics: Human Language
  Technologies, Volume 1 (Long Papers)}, pages 1875--1885, New Orleans,
  Louisiana. Association for Computational Linguistics.

\bibitem[{Kazemnejad et~al.(2020)Kazemnejad, Salehi, and
  Baghshah}]{kazemnejad2020paraphrase}
Amirhossein Kazemnejad, Mohammadreza Salehi, and Mahdieh~Soleymani Baghshah.
  2020.
\newblock Paraphrase generation by learning how to edit from samples.
\newblock In \emph{Proceedings of the 58th Annual Meeting of the Association
  for Computational Linguistics}, pages 6010--6021.

\bibitem[{Kim and Rush(2016)}]{kim-rush-2016-sequence}
Yoon Kim and Alexander~M. Rush. 2016.
\newblock \href {https://doi.org/10.18653/v1/D16-1139} {Sequence-level
  knowledge distillation}.
\newblock In \emph{Proceedings of the 2016 Conference on Empirical Methods in
  Natural Language Processing}, pages 1317--1327, Austin, Texas. Association
  for Computational Linguistics.

\bibitem[{Kumar et~al.(2019)Kumar, Bhattamishra, Bhandari, and
  Talukdar}]{kumar-etal-2019-submodular}
Ashutosh Kumar, Satwik Bhattamishra, Manik Bhandari, and Partha Talukdar. 2019.
\newblock \href {https://doi.org/10.18653/v1/N19-1363} {Submodular
  optimization-based diverse paraphrasing and its effectiveness in data
  augmentation}.
\newblock In \emph{Proceedings of the 2019 Conference of the North {A}merican
  Chapter of the Association for Computational Linguistics: Human Language
  Technologies, Volume 1 (Long and Short Papers)}, pages 3609--3619,
  Minneapolis, Minnesota. Association for Computational Linguistics.

\bibitem[{Lan et~al.(2017)Lan, Qiu, He, and Xu}]{lan-etal-2017-continuously}
Wuwei Lan, Siyu Qiu, Hua He, and Wei Xu. 2017.
\newblock \href {https://doi.org/10.18653/v1/D17-1126} {A continuously growing
  dataset of sentential paraphrases}.
\newblock In \emph{Proceedings of the 2017 Conference on Empirical Methods in
  Natural Language Processing}, pages 1224--1234, Copenhagen, Denmark.
  Association for Computational Linguistics.

\bibitem[{Lewis et~al.(2019)Lewis, Liu, Goyal, Ghazvininejad, Mohamed, Levy,
  Stoyanov, and Zettlemoyer}]{lewis2019bart}
Mike Lewis, Yinhan Liu, Naman Goyal, Marjan Ghazvininejad, Abdelrahman Mohamed,
  Omer Levy, Ves Stoyanov, and Luke Zettlemoyer. 2019.
\newblock Bart: Denoising sequence-to-sequence pre-training for natural
  language generation, translation, and comprehension.
\newblock \emph{arXiv preprint arXiv:1910.13461}.

\bibitem[{Lewis et~al.(2020)Lewis, Liu, Goyal, Ghazvininejad, Mohamed, Levy,
  Stoyanov, and Zettlemoyer}]{lewis-etal-2020-bart}
Mike Lewis, Yinhan Liu, Naman Goyal, Marjan Ghazvininejad, Abdelrahman Mohamed,
  Omer Levy, Veselin Stoyanov, and Luke Zettlemoyer. 2020.
\newblock \href {https://doi.org/10.18653/v1/2020.acl-main.703} {{BART}:
  Denoising sequence-to-sequence pre-training for natural language generation,
  translation, and comprehension}.
\newblock In \emph{Proceedings of the 58th Annual Meeting of the Association
  for Computational Linguistics}, pages 7871--7880, Online. Association for
  Computational Linguistics.

\bibitem[{Li et~al.(2017)Li, Jiang, Shang, and Li}]{li2017paraphrase}
Zichao Li, Xin Jiang, Lifeng Shang, and Hang Li. 2017.
\newblock Paraphrase generation with deep reinforcement learning.
\newblock \emph{arXiv preprint arXiv:1711.00279}.

\bibitem[{Li et~al.(2019)Li, Jiang, Shang, and Liu}]{li-etal-2019-decomposable}
Zichao Li, Xin Jiang, Lifeng Shang, and Qun Liu. 2019.
\newblock \href {https://doi.org/10.18653/v1/P19-1332} {Decomposable neural
  paraphrase generation}.
\newblock In \emph{Proceedings of the 57th Annual Meeting of the Association
  for Computational Linguistics}, pages 3403--3414, Florence, Italy.
  Association for Computational Linguistics.

\bibitem[{Lin(2004)}]{lin2004rouge}
Chin-Yew Lin. 2004.
\newblock Rouge: A package for automatic evaluation of summaries.
\newblock In \emph{Text summarization branches out}, pages 74--81.

\bibitem[{Lin et~al.(2014)Lin, Maire, Belongie, Hays, Perona, Ramanan,
  Doll{\'a}r, and Zitnick}]{lin2014microsoft}
Tsung-Yi Lin, Michael Maire, Serge Belongie, James Hays, Pietro Perona, Deva
  Ramanan, Piotr Doll{\'a}r, and C~Lawrence Zitnick. 2014.
\newblock Microsoft coco: Common objects in context.
\newblock In \emph{European conference on computer vision}, pages 740--755.
  Springer.

\bibitem[{Liu et~al.(2020)Liu, Mou, Meng, Zhou, Zhou, and
  Song}]{liu-etal-2020-unsupervised}
Xianggen Liu, Lili Mou, Fandong Meng, Hao Zhou, Jie Zhou, and Sen Song. 2020.
\newblock \href {https://doi.org/10.18653/v1/2020.acl-main.28} {Unsupervised
  paraphrasing by simulated annealing}.
\newblock In \emph{Proceedings of the 58th Annual Meeting of the Association
  for Computational Linguistics}, pages 302--312, Online. Association for
  Computational Linguistics.

\bibitem[{Mao and Lee(2019)}]{mao-lee-2019-polly}
Hong-Ren Mao and Hung-Yi Lee. 2019.
\newblock \href {https://doi.org/10.18653/v1/D19-1611} {Polly want a cracker:
  Analyzing performance of parroting on paraphrase generation datasets}.
\newblock In \emph{Proceedings of the 2019 Conference on Empirical Methods in
  Natural Language Processing and the 9th International Joint Conference on
  Natural Language Processing (EMNLP-IJCNLP)}, pages 5960--5968, Hong Kong,
  China. Association for Computational Linguistics.

\bibitem[{McKeown(1980)}]{mckeown1980paraphrasing}
Kathleen~R McKeown. 1980.
\newblock Paraphrasing using given and new information in a question-answer
  system.
\newblock \emph{Technical Reports (CIS)}, page 723.

\bibitem[{Meteer and Shaked(1988)}]{meteer1988strategies}
Marie Meteer and Varda Shaked. 1988.
\newblock Strategies for effective paraphrasing.
\newblock In \emph{Coling Budapest 1988 Volume 2: International Conference on
  Computational Linguistics}.

\bibitem[{Miao et~al.(2018)Miao, Zhou, Mou, Yan, and Li}]{miao2018cgmh}
Ning Miao, Hao Zhou, Lili Mou, Rui Yan, and Lei Li. 2018.
\newblock \href {http://arxiv.org/abs/1811.10996} {Cgmh: Constrained sentence
  generation by metropolis-hastings sampling}.

\bibitem[{Mikolov et~al.(2013)Mikolov, Chen, Corrado, and
  Dean}]{mikolov2013efficient}
Tomas Mikolov, Kai Chen, Greg Corrado, and Jeffrey Dean. 2013.
\newblock Efficient estimation of word representations in vector space.
\newblock \emph{arXiv preprint arXiv:1301.3781}.

\bibitem[{Miller(1998)}]{miller1998wordnet}
George~A Miller. 1998.
\newblock \emph{WordNet: An electronic lexical database}.
\newblock MIT press.

\bibitem[{Moon et~al.(2019)Moon, Awasthy, Ni, and Florian}]{moon2019towards}
Taesun Moon, Parul Awasthy, Jian Ni, and Radu Florian. 2019.
\newblock Towards lingua franca named entity recognition with bert.
\newblock \emph{arXiv preprint arXiv:1912.01389}.

\bibitem[{Papineni et~al.(2002)Papineni, Roukos, Ward, and
  Zhu}]{papineni-etal-2002-bleu}
Kishore Papineni, Salim Roukos, Todd Ward, and Wei-Jing Zhu. 2002.
\newblock \href {https://doi.org/10.3115/1073083.1073135} {{B}leu: a method for
  automatic evaluation of machine translation}.
\newblock In \emph{Proceedings of the 40th Annual Meeting of the Association
  for Computational Linguistics}, pages 311--318, Philadelphia, Pennsylvania,
  USA. Association for Computational Linguistics.

\bibitem[{Prakash et~al.(2016)Prakash, Hasan, Lee, Datla, Qadir, Liu, and
  Farri}]{prakash-etal-2016-neural}
Aaditya Prakash, Sadid~A. Hasan, Kathy Lee, Vivek Datla, Ashequl Qadir, Joey
  Liu, and Oladimeji Farri. 2016.
\newblock \href {https://www.aclweb.org/anthology/C16-1275} {Neural paraphrase
  generation with stacked residual {LSTM} networks}.
\newblock In \emph{Proceedings of {COLING} 2016, the 26th International
  Conference on Computational Linguistics: Technical Papers}, pages 2923--2934,
  Osaka, Japan. The COLING 2016 Organizing Committee.

\bibitem[{Qian et~al.(2019)Qian, Qiu, Zhang, Jiang, and Yu}]{qian2019exploring}
Lihua Qian, Lin Qiu, Weinan Zhang, Xin Jiang, and Yong Yu. 2019.
\newblock Exploring diverse expressions for paraphrase generation.
\newblock In \emph{Proceedings of the 2019 Conference on Empirical Methods in
  Natural Language Processing and the 9th International Joint Conference on
  Natural Language Processing (EMNLP-IJCNLP)}.

\bibitem[{Raffel et~al.(2019)Raffel, Shazeer, Roberts, Lee, Narang, Matena,
  Zhou, Li, and Liu}]{raffel2019exploring}
Colin Raffel, Noam Shazeer, Adam Roberts, Katherine Lee, Sharan Narang, Michael
  Matena, Yanqi Zhou, Wei Li, and Peter~J Liu. 2019.
\newblock Exploring the limits of transfer learning with a unified text-to-text
  transformer.
\newblock \emph{arXiv preprint arXiv:1910.10683}.

\bibitem[{Reimers and Gurevych(2019)}]{reimers-gurevych-2019-sentence}
Nils Reimers and Iryna Gurevych. 2019.
\newblock \href {https://doi.org/10.18653/v1/D19-1410} {Sentence-{BERT}:
  Sentence embeddings using {S}iamese {BERT}-networks}.
\newblock In \emph{Proceedings of the 2019 Conference on Empirical Methods in
  Natural Language Processing and the 9th International Joint Conference on
  Natural Language Processing (EMNLP-IJCNLP)}, pages 3982--3992, Hong Kong,
  China. Association for Computational Linguistics.

\bibitem[{Roy and Grangier(2019)}]{roy-grangier-2019-unsupervised}
Aurko Roy and David Grangier. 2019.
\newblock \href {https://doi.org/10.18653/v1/P19-1605} {Unsupervised
  paraphrasing without translation}.
\newblock In \emph{Proceedings of the 57th Annual Meeting of the Association
  for Computational Linguistics}, pages 6033--6039, Florence, Italy.
  Association for Computational Linguistics.

\bibitem[{See et~al.(2017)See, Liu, and Manning}]{see2017pointergenerator}
Abigail See, Peter~J. Liu, and Christopher~D. Manning. 2017.
\newblock Get to the point: Summarization with pointer-generator networks.
\newblock In \emph{Proceedings of the 55th Annual Meeting of the Association
  for Computational Linguistics (Volume 1: Long Papers)}.

\bibitem[{Sellam et~al.(2020)Sellam, Das, and Parikh}]{sellam-etal-2020-bleurt}
Thibault Sellam, Dipanjan Das, and Ankur Parikh. 2020.
\newblock \href {https://doi.org/10.18653/v1/2020.acl-main.704} {{BLEURT}:
  Learning robust metrics for text generation}.
\newblock In \emph{Proceedings of the 58th Annual Meeting of the Association
  for Computational Linguistics}, pages 7881--7892, Online. Association for
  Computational Linguistics.

\bibitem[{Shimanaka et~al.(2018)Shimanaka, Kajiwara, and
  Komachi}]{shimanaka-etal-2018-ruse}
Hiroki Shimanaka, Tomoyuki Kajiwara, and Mamoru Komachi. 2018.
\newblock \href {https://doi.org/10.18653/v1/W18-6456} {{RUSE}: Regressor using
  sentence embeddings for automatic machine translation evaluation}.
\newblock In \emph{Proceedings of the Third Conference on Machine Translation:
  Shared Task Papers}, pages 751--758, Belgium, Brussels. Association for
  Computational Linguistics.

\bibitem[{Siddhant et~al.(2020)Siddhant, Bapna, Cao, Firat, Chen, Kudugunta,
  Arivazhagan, and Wu}]{siddhant-etal-2020-leveraging}
Aditya Siddhant, Ankur Bapna, Yuan Cao, Orhan Firat, Mia Chen, Sneha Kudugunta,
  Naveen Arivazhagan, and Yonghui Wu. 2020.
\newblock \href {https://doi.org/10.18653/v1/2020.acl-main.252} {Leveraging
  monolingual data with self-supervision for multilingual neural machine
  translation}.
\newblock In \emph{Proceedings of the 58th Annual Meeting of the Association
  for Computational Linguistics}, pages 2827--2835, Online. Association for
  Computational Linguistics.

\bibitem[{Siddique et~al.(2020)Siddique, Oymak, and
  Hristidis}]{siddique2020unsupervised}
AB~Siddique, Samet Oymak, and Vagelis Hristidis. 2020.
\newblock Unsupervised paraphrasing via deep reinforcement learning.
\newblock In \emph{Proceedings of the 26th ACM SIGKDD International Conference
  on Knowledge Discovery \& Data Mining}, pages 1800--1809.

\bibitem[{Sun and Zhou(2012)}]{sun-zhou-2012-joint}
Hong Sun and Ming Zhou. 2012.
\newblock \href {https://www.aclweb.org/anthology/P12-2008} {Joint learning of
  a dual {SMT} system for paraphrase generation}.
\newblock In \emph{Proceedings of the 50th Annual Meeting of the Association
  for Computational Linguistics (Volume 2: Short Papers)}, pages 38--42, Jeju
  Island, Korea. Association for Computational Linguistics.

\bibitem[{Thompson and Post(2020)}]{thompson-post-2020-automatic}
Brian Thompson and Matt Post. 2020.
\newblock \href {https://doi.org/10.18653/v1/2020.emnlp-main.8} {Automatic
  machine translation evaluation in many languages via zero-shot paraphrasing}.
\newblock In \emph{Proceedings of the 2020 Conference on Empirical Methods in
  Natural Language Processing (EMNLP)}, pages 90--121, Online. Association for
  Computational Linguistics.

\bibitem[{Vaswani et~al.(2017)Vaswani, Shazeer, Parmar, Uszkoreit, Jones,
  Gomez, Kaiser, and Polosukhin}]{vaswani2017attention}
Ashish Vaswani, Noam Shazeer, Niki Parmar, Jakob Uszkoreit, Llion Jones,
  Aidan~N. Gomez, Lukasz Kaiser, and Illia Polosukhin. 2017.
\newblock \href {http://arxiv.org/abs/1706.03762} {Attention is all you need}.

\bibitem[{Vijayakumar et~al.(2018)Vijayakumar, Cogswell, Selvaraju, Sun, Lee,
  Crandall, and Batra}]{vijayakumar2018diverse}
Ashwin Vijayakumar, Michael Cogswell, Ramprasaath Selvaraju, Qing Sun, Stefan
  Lee, David Crandall, and Dhruv Batra. 2018.
\newblock Diverse beam search for improved description of complex scenes.
\newblock In \emph{Proceedings of the AAAI Conference on Artificial
  Intelligence}, volume~32.

\bibitem[{Wieting and Gimpel(2018)}]{wieting-gimpel-2018-paranmt}
John Wieting and Kevin Gimpel. 2018.
\newblock \href {https://doi.org/10.18653/v1/P18-1042} {{P}ara{NMT}-50{M}:
  Pushing the limits of paraphrastic sentence embeddings with millions of
  machine translations}.
\newblock In \emph{Proceedings of the 56th Annual Meeting of the Association
  for Computational Linguistics (Volume 1: Long Papers)}, pages 451--462,
  Melbourne, Australia. Association for Computational Linguistics.

\bibitem[{Witteveen and Andrews(2019)}]{witteveen2019paraphrasing}
Sam Witteveen and Martin Andrews. 2019.
\newblock Paraphrasing with large language models.
\newblock \emph{arXiv preprint arXiv:1911.09661}.

\bibitem[{Xu et~al.(2013)Xu, Ritter, and Grishman}]{xu2013gathering}
Wei Xu, Alan Ritter, and Ralph Grishman. 2013.
\newblock Gathering and generating paraphrases from twitter with application to
  normalization.
\newblock In \emph{Proceedings of the sixth workshop on building and using
  comparable corpora}, pages 121--128.

\bibitem[{Xue et~al.(2020)Xue, Constant, Roberts, Kale, Al-Rfou, Siddhant,
  Barua, and Raffel}]{xue2020mt5}
Linting Xue, Noah Constant, Adam Roberts, Mihir Kale, Rami Al-Rfou, Aditya
  Siddhant, Aditya Barua, and Colin Raffel. 2020.
\newblock mt5: A massively multilingual pre-trained text-to-text transformer.
\newblock \emph{arXiv preprint arXiv:2010.11934}.

\bibitem[{Yang et~al.(2019)Yang, Dai, Yang, Carbonell, Salakhutdinov, and
  Le}]{yang2019xlnet}
Zhilin Yang, Zihang Dai, Yiming Yang, Jaime Carbonell, Russ~R Salakhutdinov,
  and Quoc~V Le. 2019.
\newblock Xlnet: Generalized autoregressive pretraining for language
  understanding.
\newblock In \emph{Advances in neural information processing systems}, pages
  5753--5763.

\bibitem[{Yin et~al.(2015)Yin, Duan, Kao, Bao, and Zhou}]{yin2015answering}
Pengcheng Yin, Nan Duan, Ben Kao, Junwei Bao, and Ming Zhou. 2015.
\newblock Answering questions with complex semantic constraints on open
  knowledge bases.
\newblock In \emph{Proceedings of the 24th ACM International on Conference on
  Information and Knowledge Management}, pages 1301--1310.

\bibitem[{Yu et~al.(2018)Yu, Dohan, Luong, Zhao, Chen, Norouzi, and
  Le}]{yu2018qanet}
Adams~Wei Yu, David Dohan, Minh-Thang Luong, Rui Zhao, Kai Chen, Mohammad
  Norouzi, and Quoc~V. Le. 2018.
\newblock {QANet: Combining Local Convolution with Global Self-Attention for
  Reading Comprehension}.
\newblock In \emph{6th International Conference on Learning Representations
  (ICLR)}.

\bibitem[{Zhang et~al.(2019)Zhang, Kishore, Wu, Weinberger, and
  Artzi}]{zhang2019bertscore}
Tianyi Zhang, Varsha Kishore, Felix Wu, Kilian~Q Weinberger, and Yoav Artzi.
  2019.
\newblock Bertscore: Evaluating text generation with bert.
\newblock \emph{arXiv preprint arXiv:1904.09675}.

\bibitem[{Zhu et~al.(2015)Zhu, Kiros, Zemel, Salakhutdinov, Urtasun, Torralba,
  and Fidler}]{Zhu_2015_ICCV}
Yukun Zhu, Ryan Kiros, Rich Zemel, Ruslan Salakhutdinov, Raquel Urtasun,
  Antonio Torralba, and Sanja Fidler. 2015.
\newblock Aligning books and movies: Towards story-like visual explanations by
  watching movies and reading books.
\newblock In \emph{The IEEE International Conference on Computer Vision
  (ICCV)}.

\end{thebibliography}
\bibliographystyle{acl_natbib}

\null\newpage

\appendix
\section{Automatic Metric Results}
\label{sec:automatic-metric-results}
We present automatic evaluation results on the previous metrics for QQP in Table~\ref{tab:qqp-old} and for ParaNMT in Table~\ref{tab:nmt-old}. We can see that for QQP our task-adaptation model without Dynamic Blocking during inference achieves state-of-the-art results among unsupervised approaches. Had we based our judgments on Table~\ref{tab:qqp-old}, we would have mistakenly selected this one as our final model.

\begin{table}[t]
\tiny
\setlength{\tabcolsep}{3pt}
  \centering
    \begin{tabular}{clcccc}
    \toprule
    \multicolumn{2}{c}{\multirow{2}[1]{*}{Model}} &
    \multicolumn{4}{c}{Quora} \\
\cmidrule{3-6}    \multicolumn{1}{c}{} & \multicolumn{1}{c}{} & \multicolumn{1}{c}{iBLEU} & \multicolumn{1}{c}{BLEU} & \multicolumn{1}{c}{ROUGE-1} & \multicolumn{1}{c}{ROUGE-2} \\
    \midrule
    \multirow{6}[0]{*}{Supervised} & ResidualLSTM & 12.67 & 17.57 & 59.22 & 32.40 \\
    \multicolumn{1}{c}{} & VAE-SVG-eq & 15.17 & 20.04 & 59.98 & 33.30 \\
    \multicolumn{1}{c}{} & Pointer-generator & 16.79 & 22.65 & 61.96 & 36.07 \\
    \multicolumn{1}{c}{} & Transformer & 16.25 & 21.73 & 60.25 & 33.45 \\
    \multicolumn{1}{c}{} & + Copy & 17.98 & 24.77 & 63.34 & 37.31 \\
    \multicolumn{1}{c}{} & DNPG  & \textbf{18.01} & \textbf{25.03} & \textbf{63.73} & \textbf{37.75} \\
    \midrule
    \multirow{5}[2]{*}{Supervised (Wiki)} & Pointer-generator & 5.04  & 6.96  & 41.89 & 12.77 \\
    \multicolumn{1}{c}{} & Transformer + Copy & 6.17  & 8.15  & 44.89 & 14.79 \\
    \multicolumn{1}{c}{} & Shallow fusion & 6.04  & 7.95  & 44.87 & 14.79 \\
    \multicolumn{1}{c}{} & Multi-task learning   & 4.90  & 6.37  & 37.64 & 11.83 \\
    \multicolumn{1}{c}{} & + Copy & 7.22  & 9.83  & 47.08 & 19.03 \\
    \multicolumn{1}{c}{} & DNPG  & \textbf{10.39} & \textbf{16.98} & \textbf{56.01} & \textbf{28.61} \\
    \midrule
    \multirow{8}[0]{*}{Unsupervised} 
        & VAE   & 8.16  & 13.96 & 44.55 & 22.64 \\
        & CGMH  & 9.94  & 15.73 & 48.73 & 26.12 \\
        & UPSA  & 12.02 & 18.18 & 56.51 & 30.69 \\
        & PUP   & 14.91 & 19.68 & 59.77 & 30.47 \\
        & \textbf{CorruptLM} & 12.08 & 16.80 & 49.13 & 26.15 \\
        & \textbf{TA} & \textbf{16.02} & \textbf{21.18} & \textbf{61.90} & \textbf{35.07} \\
        & \textbf{TA+SS} & 15.57 & 20.68 & 61.51 & 32.78 \\
        & \textbf{TA+SS+DB} & 9.67 & 14.12 & 60.06 & 28.91 \\
    \midrule
    No model & \textbf{Copy-input} & \textbf{24.79} & \textbf{30.98} & \textbf{65.60} & \textbf{42.09} \\
    \bottomrule
    \end{tabular}%
  \caption{Automatic evaluation results on the QQP dataset. Models we (re)produced and SOTA results in each category are boldfaced. "Supervised (Wiki)" stands for models trained on WikiAnswers and evaluated on QQP.}
  \label{tab:qqp-old}%
\end{table}%

\begin{table}[t]
\tiny
  \centering
    \begin{tabular}{clcccc}
    \toprule
    \multicolumn{2}{c}{\multirow{2}[1]{*}{Model}} & \multicolumn{4}{c}{Oracle Quality (10 sentences)} \\
    \cmidrule{3-6}
    \multicolumn{2}{c}{} & BLEU  & ROUGE-1 & ROUGE-2 & ROUGE-L \\
    \midrule
    \multirow{6}[0]{*}{Supervised} 
          & copy-input & 18.4  & 54.4  & 27.2  & 49.2 \\
          & SCPN  & 21.3  & 53.2  & 30.3  & 51.0 \\
          & Transformer seq2seq & \textbf{32.8}  & \textbf{63.1}  & \textbf{41.4}  & \textbf{63.3} \\
          & + diverse-decoding & 24.8  & 56.8  & 33.2  & 56.4 \\
          & SOW-REAP (LSTM) & 27.0  & 57.9  & 34.8  & 57.5 \\
          & SOW-REAP & 30.9  & 62.3 & 40.2 & 61.7 \\
    \midrule
    \multirow{3}[0]{*}{Unsupervised}
        & \textbf{CorruptLM (QQP)} & 7.6 & 31.9 & 11.6 & 31.6 \\
        & \textbf{TA+SS+DB (QQP)} & 22.0 & \textbf{60.1} & \textbf{33.8}  & \textbf{54.9} \\
        & \textbf{TA+SS+DB} & \textbf{22.6}  & 59.8  & 33.2  & 54.5 \\
    \bottomrule
    \end{tabular}%
  \caption{Automatic metrics results on the Para-NMT dataset. "(QQP)" stands for models finetuned on the non-parallel QQP dataset and evaluated on the ParaNMT dataset.}
  \label{tab:nmt-old}%
\end{table}%

\section{Robustness to Grammar Errors}
\label{sec:robustness-to-grammar-errors}
During the task-adaptation phase, our model in most cases has a grammatically correct sentence as the target sequence. Additionally, shuffling during that phase encourages the model to attend to the context during generation. These setups make our model reasonably robust to grammar errors so that it can paraphrase and normalize the input at the same time. Table~\ref{tab:grammar} shows a case where we intentionally introduce grammar errors on subject-verb agreement, singular vs. plural, and verb inflections. We find that our model is in most cases robust to such errors. This trait is desired because we may face noisy inputs from users. Through early ablation studies, we observed that without shuffling during task-adaptation, the model was much less robust to grammar errors. Hence shuffling does more than just improving on the BERT-\textit{i}BLEU metric (Table~\ref{tab:ablation}). 

\begin{table}[t]
\tiny
  \centering
    \begin{tabular}{cl}
    \toprule
    \multicolumn{1}{c}{Input} & Our approach \underline{\smash{are}} data-driven and can be \underline{\smash{apply}} across various \underline{\smash{situation}}. \\
    \midrule
    \multirow{8}[0]{*}{Output} & Our approach is data-driven and can be applied across various situations. \\
        & Our approach is data-driven and can be applied across different situations. \\
        & Our approach is data-driven and can be applied across diverse situations. \\
        & Our approaches are data-driven and can be applied across various situations. \\
        & Our data-driven approach can be applied across different situations. \\
        & Our approaches are data-driven and can be applied across different situations. \\
        & Our data-driven approach can be applied across diverse situations. \\
        & Our approaches are data-driven and can be applied across diverse situations. \\
    \bottomrule
    \end{tabular}%
  \caption{Selected example of output candidates produced by our model where we intentionally introduce grammar errors (marked with underlines). We observe that all paraphrase candidates have these errors corrected.}
  \label{tab:grammar}%
\end{table}%

\section{Failure Modes}
\label{sec:failure-modes}
Though only occurring occasionally, our model exhibits multiple failure patterns. Hence we perform "anti-cherry-picking" and present in Table~\ref{tab:failure} some of such examples and the respective modes we outline. We hypothesize that the Antonym mode can be partially addressed by a lookup in the dictionary to additionally block the antonyms. Grammar errors are harder to resolve because they are usually apparent only after the whole sentence is generated. A grammar checker on the candidates may improve the situation. The swapping of subject and object shows that unsupervised approaches based on pre-trained language models could only carry us so far till the syntactic-level. In its current form, it cannot handle semantic mistakes. For missing named entities, an NER tagger can help filter candidates that miss important entities. We leave addressing these failure modes as future work.

\begin{table*}[t]
\small
  \centering
    \begin{tabular}{lll}
    \toprule
    \multicolumn{1}{c}{Failure mode} & \multicolumn{1}{c}{Input} & \multicolumn{1}{c}{Output}\\
    \midrule
    Antonym & How do I \underline{gain} weight in a healthy way? & How do I \underline{lose} weight in healthy ways? \\
    Repeated words & What is the \underline{funniest} movie to watch?  &  What is the \underline{most} \underline{funniest} film to see? \\
    Grammar errors & Do spirits or ghosts exist?  &    \underline{Do} ghost or spirit exist? \\
    Subject $\leftrightarrow$ object & How will you know \underline{you} love \underline{someone}?   &  How will you tell if \underline{someone} loves \underline{you}? \\ 
    Missing named entity & A look of dismay came into \underline{luzhin}'s face.  &     A look of disappointment came into the face. \\
    \bottomrule
    \end{tabular}%
  \caption{Typical examples where our model fails to generate correct paraphrases. Words related to each failure mode are underlined.}
  \label{tab:failure}%
\end{table*}%

\section{Details of Dynamic Blocking}
\label{sec:details-of-dynamic-blocking}
\paragraph{Block surface-form variations}
In our early experiments, we observed that when blocking a word (e.g. "give"), the model usually tries to generate its capitalized ("Give") or upper ("GIVE") version. From we human's perspective, these are usually not good paraphrases -- intuitively we would prefer a different word. Similar to whole-word masking introduced in later versions of BERT,\footnote{\url{https://github.com/google-research/bert}} we only block the beginning of the word rather than any subword.

\paragraph{Block Closed-Class Words}
We also leverage linguistic knowledge to help boost the quality of the paraphrases by avoiding blocking closed-class words, or functional words.\footnote{\url{https://mailman.uib.no/public/corpora/attachments/20111124/6c58cb02/attachment.txt}}
The closed classes in English include pronouns, determiners, conjunctions, and prepositions while open-class words correspond to nouns, lexical verbs, adjectives, and adverbs. There are two justifications for blocking these words. First, because they are closed-class, there are fewer synonyms available; second, blocking such words is error-prone. For example, changing determiners (e.g. from "\textit{you}" to "\textit{I}") may lead to syntactic or semantic errors, while modifying conjunctions (e.g. from "\textit{and}" to "\textit{or}") may lead to change in logical relationships.

\paragraph{Block Inflections}
\label{sec:block-inflections}
In Section~\ref{para:dynamic-blocking-without-finetuning}, we mentioned that BART can directly work with Dynamic Blocking without task-adaptation or self-supervision, but that results in lower quality, especially lacking syntactic variance because it is not trained with the shuffling strategy during task-adaptation. In addition, we found that without finetuning, BART tries to generate inflections of a word when it is blocked. To partially remedy this drawback, we use the \textit{pattern} library\footnote{\url{https://github.com/clips/pattern}} to enumerate all inflections of a word to block (e.g. for "\textit{give}" we should also block "\textit{gives}", "\textit{gave}", "\textit{giving}" and "\textit{given}") in addition to all the other blocking schemes introduced in Section~\ref{sec:exp-setup}.  This is available for most languages that involve inflections. We show in Table~\ref{tab:inflections} the output candidates of a selected example with and without blocking inflections.

\paragraph{Retain Named Entities}
We also explore a variation of the system where we employ a separate Named Entity Recognition model to identify the named entities in the source sequence and prevent any tokens in these entities from appearing in the full block dictionary. This change ensures that all named entities are copied verbatim.

\begin{table*}[t]
\small
  \centering
    \begin{tabular}{c|c|l}
    \toprule
    \multicolumn{2}{c|}{Input} & The random selection of pages must be performed by someone other than the player. \\
    \midrule
    \multirow{10}[0]{*}{Output} & \multirow{5}[2]{*}{Blocking inflections} & The random choice of the pages must be performed by someone else than the player. \\
    & & The random selection of the pages must be performed by someone else than the user. \\
    & & The random selection of the pages must be executed by someone other than the user. \\
    & & The random collection of these pages must be performed by someone else than the player. \\
    & & The random selection of these pages must be executed by someone other than the user. \\
    \cmidrule{2-3}          
    & \multirow{5}[0]{*}{No blocking inflections} & The \underline{randomly} selection of \underline{page} must be \underline{perform} by someone else than the players. \\
    & & The random choice of \underline{page} must be performed by someone else than the player. \\
    & & The \underline{randomly} selection of \underline{page} must be \underline{perform} by someone rather than the players. \\
    & & The random choice of \underline{page} must be performed by someone rather than the player. \\
    & & The random collection of pages must be performed by someone else than the players. \\
    \bottomrule
    \end{tabular}%
  \caption{Selected example of output candidates produced by BART + Dynamic Blocking with and without blocking inflections. Grammar errors made by the latter due to wrong inflections are underlined.}
  \label{tab:inflections}%
\end{table*}%

\section{Paraphrasing in German}
\label{sec:paraphrasing-in-german}
We pair BART directly with Dynamic Blocking to generate paraphrases in German. In Table~\ref{tab:de}, we can see that all candidates (left column) have different surface forms, while all translations in English (right column)\footnote{By Google Translator: \url{https://translate.google.com/}} share similar meanings.

\begin{table*}[t]
\small
  \centering
    \begin{tabular}{cll}
    \toprule
          & \multicolumn{1}{c}{German} & \multicolumn{1}{c}{Translation from German} \\
    \midrule
    Input & Warum finden keine Brandschutzbelehrungen statt ? & Why are there no fire instructions? \\
    \midrule
    \multirow{5}[0]{*}{Candidates} 
        & Warum lieen keine Geschutzbelehrungen statt? & Why were there no protection instructions? \\
        & Warum finden keine Geschutzbelehrungen statt? & Why are there no protection instructions? \\
        & Warum lieen keine Brandschutzbelehrungen statt? & Why weren't there any fire safety instructions? \\
        & Warum finden keine Geschutzbelehrungen statt? & Why are there no protection instructions? \\
        & Warum finden wir keine Brandschutzbelehrungen statt? & Why are we not giving fire safety instructions? \\
    \bottomrule
    \end{tabular}%
  \caption{Paraphrasing German input by directly applying Dynamic Blocking to BART. Translations on the right are given by the Google Translator, except that the first one is the ground-truth translation. Note that the candidates are ranked by multi-lingual BERT rather than RoBERTa-base which is only used to rank English outputs.}
  \label{tab:de}%
\end{table*}%

\section{MTurk Instructions}
\label{sec:mturk instructions}
\begin{figure*}[t]
\centering
\includegraphics[width=1.0\textwidth]{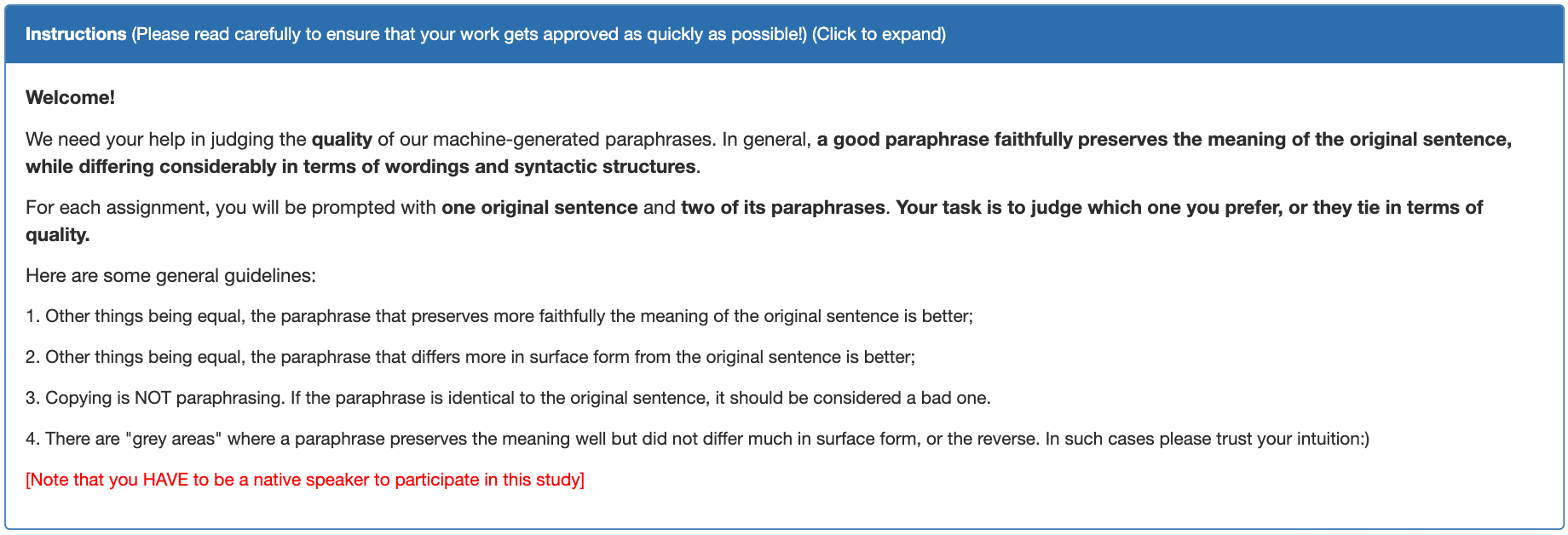}
\caption{Interface of our MTurk studies for head-to-head comparisions with other models.}
\label{fig:mturk}
\end{figure*}

\begin{figure*}[t]
\centering
\includegraphics[width=1.0\textwidth]{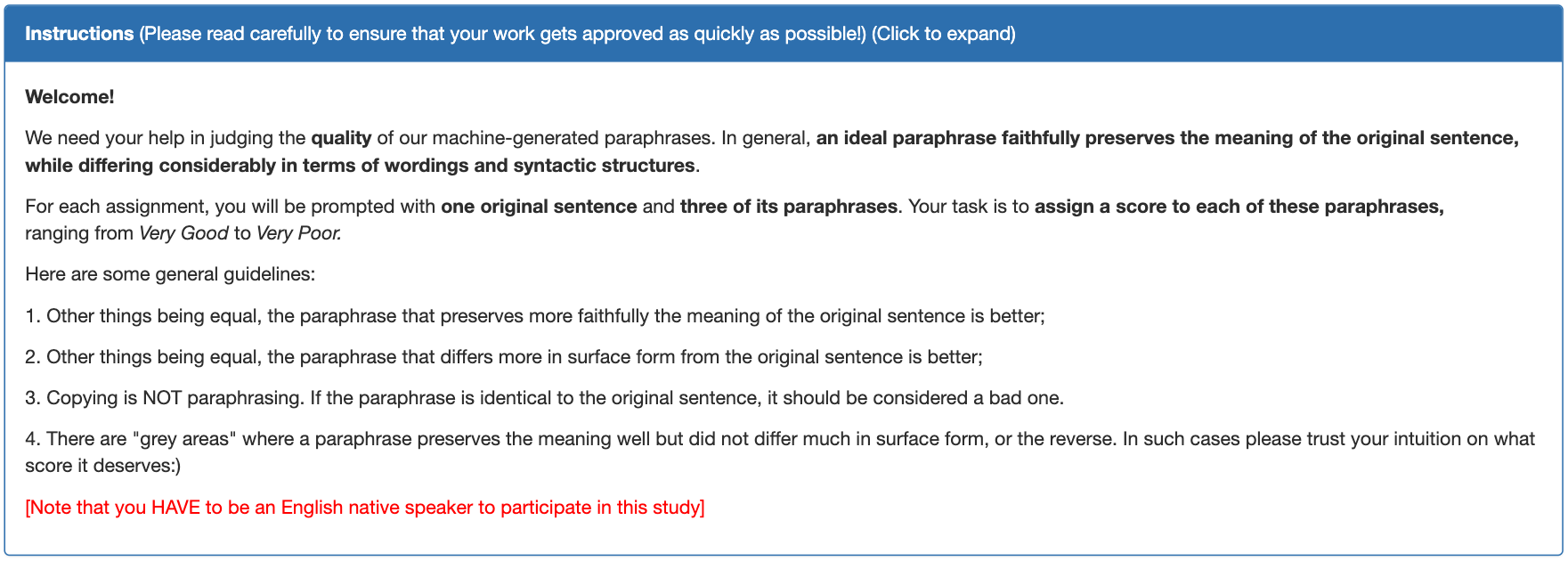}
\caption{Interface of our MTurk studies for head-to-head comparisions with other models.}
\label{fig:mturk-likert}
\end{figure*}

To facilitate reproducibility, we include our MTurk instructions for the head-to-head and the Likert-based human studies (Figure~\ref{fig:mturk} and~\ref{fig:mturk-likert}). As mentioned in Section~\ref{subsec:human-evaluation}, we only provide guidelines on which paraphrases are better in general and leave the rest to the annotator's intuition.

\end{document}